\documentclass[conference]{IEEEtran}
\IEEEoverridecommandlockouts
\usepackage{cite}
\usepackage{amsmath,amssymb,amsfonts}
\usepackage{algorithmic}
\usepackage[colorlinks,linkcolor=red,anchorcolor=blue,citecolor=green]{hyperref}
\usepackage{booktabs}
\usepackage{multirow}
\usepackage{graphicx}

\usepackage{textcomp}
\usepackage{xcolor}
\def\BibTeX{{\rm B\kern-.05em{\sc i\kern-.025em b}\kern-.08em
    T\kern-.1667em\lower.7ex\hbox{E}\kern-.125emX}}
\begin{document}

\title{FaceSkin: A Privacy Preserving Facial skin patch Dataset for multi Attributes classification
}


\author{Qiushi Guo,Shisha Liao\\
AI Lab, China Merchants Bank\\
{\tt\small guoqiushi910@cmbchina.com,liaoss@cmbchina.com}}

\maketitle

\begin{abstract}
Human facial skin images contain abundant textural information that can serve as valuable features for attribute classification, such as age, race, and gender. Additionally, facial skin images offer the advantages of easy collection and minimal privacy concerns. However, the availability of well-labeled human skin datasets with a sufficient number of images is limited. To address this issue, we introduce a dataset called FaceSkin, which encompasses a diverse range of ages and races. Furthermore, to broaden the application scenarios, we incorporate synthetic skin-patches obtained from 2D and 3D attack images, including printed paper, replays, and 3D masks. We evaluate the FaceSkin dataset across distinct categories and present experimental results demonstrating its effectiveness in attribute classification, as well as its potential for various downstream tasks, such as Face anti-spoofing and Age estimation.
 FaceSkin can be downloaded \href{https://drive.google.com/file/d/1V4eCXlH7NXxsGxiw27o9FWlvYsx2IL8i/view?usp=share_link}{here.}
\end{abstract}


\section{Introduction}
The rise of deep learning has led to a significant demand for labeled data, particularly in the context of human face images. Training convolutional neural networks (CNNs) typically requires a large number of images. However, in recent times, public awareness of privacy concerns has increased, largely driven by the advancements in social media. Collecting face images is a resource-intensive and time-consuming process. In light of these challenges, there is a need to propose a novel dataset that effectively balances the preservation of biometric properties while ensuring privacy protection. Such a dataset would address the demand for labeled data while mitigating privacy concerns in an efficient manner.
\begin{figure}[htb]
    \centering
    \includegraphics[width=0.48\textwidth]{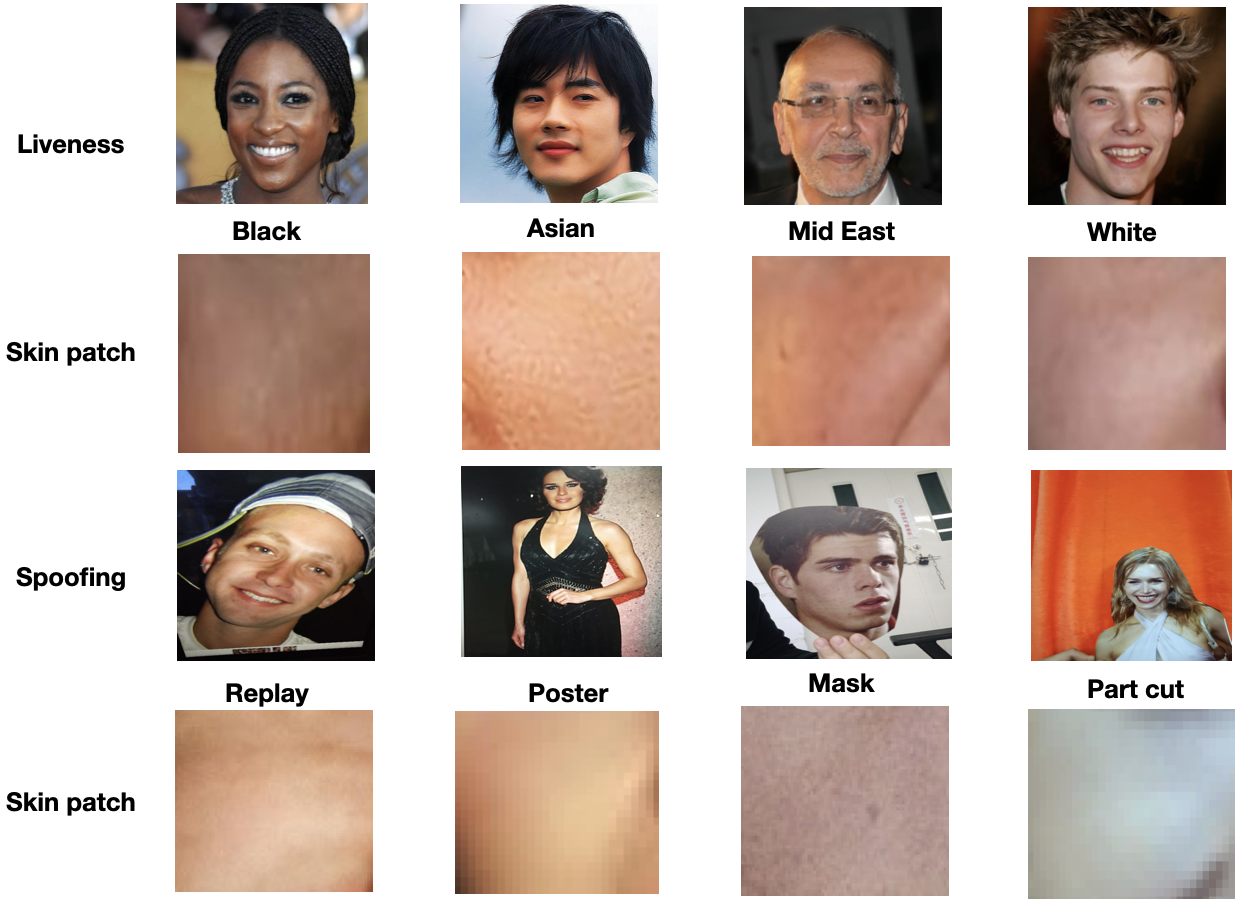 }
    \caption{sample images of FaceSkin}
    \label{fig:sample images}
\end{figure}

Prior studies\cite{atoum2017face-patch}   have demonstrated the potential of utilizing facial skin for disease detection and spoofing detection, highlighting the representational capacity of facial skin images. Drawing inspiration from these findings\cite{atoum2017face-patch} , we hypothesize that facial skin patches can be employed as a substitute for complete face images to predict attributes in various scenarios. However, constructing a qualified dataset for deep learning classification tasks necessitates satisfying certain criteria, including a large quantity of images, high quality, and a diverse range of classes. As of our knowledge, existing skin-related datasets suffer from either an inadequate number of images or a limited range of annotations, thereby impeding further exploration in this domain.


This paper introduces FaceSkin, a comprehensive dataset that encompasses diverse race and age groups. To enhance its usability across different scenarios, skin patches are extracted from synthetic faces. The Celeb-HQ dataset\cite{celeba} is selected as the primary source of original images, with non-frontal images being filtered out to ensure face image normalization. To extract pure skin patches, facial landmarks are obtained using Mediapipe\cite{lugaresi2019mediapipe}, enabling the localization and extraction of skin patches. The aforementioned patches are annotated using the approach proposed in \cite{deepface} to infer corresponding attributes such as age, race, and gender.

Given that the categories in Celeb-HQ are unevenly distributed, with a significant portion representing white individuals, the dataset is supplemented with additional face images from other races to address this imbalance. For the inclusion of spoofing face images, Celeb-spoof\cite{celeba-spoof} and 3D masks serve as the sources of original face images. Finally, each image in the dataset is annotated with the following attributes: fake/real, male/female, age, and race. To evaluate the effectiveness of FaceSkin, the dataset is benchmarked based on individual properties, namely age, gender, race, and liveness.
Additionally, we performed experiments to assess the representational capability of facial skin patches. The results demonstrate that pure facial skin patches can serve as a privacy-friendly representation of the face.
\begin{figure*}[htb]
    \centering
    \includegraphics[width=0.9\textwidth]{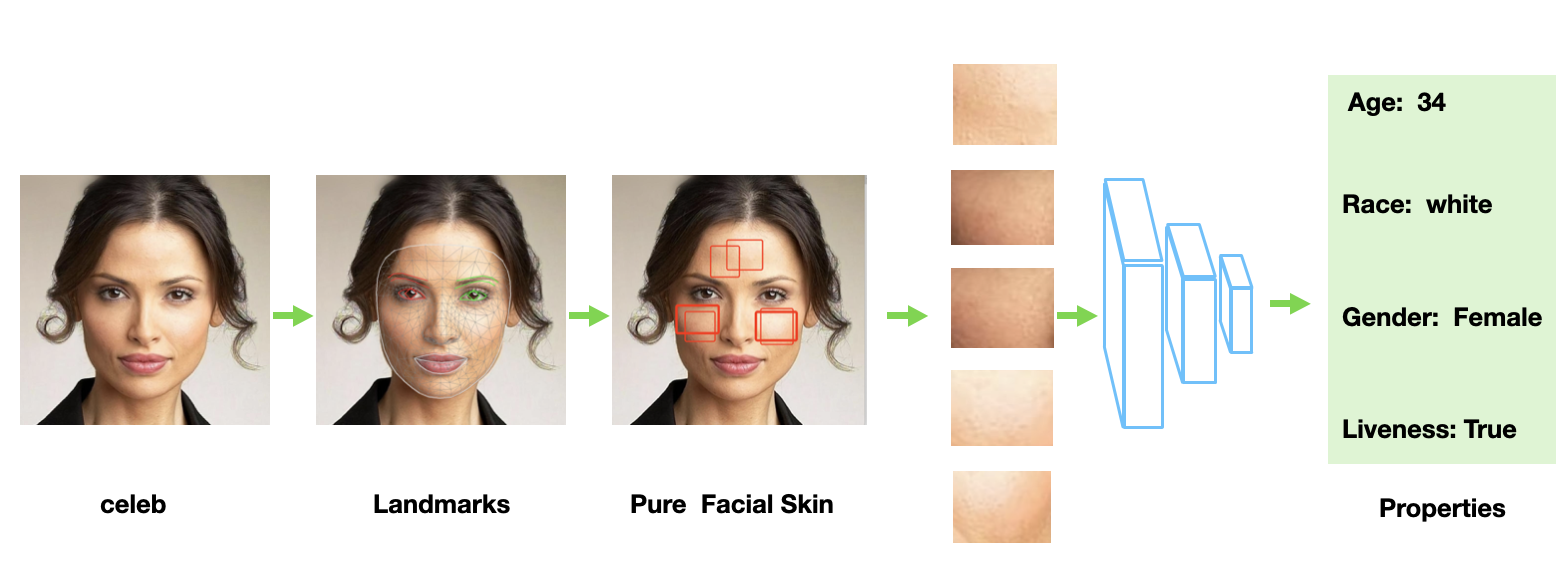 }
    \caption{Workflow of data acquisition. Firstly, landmarks of face are obtained guided by Mediapipe to locate pure skin patches. Secondly, attributes(age, race, liveness and gender) are obtained by feeding the images to DeepFace\cite{deepface}. Finally, the labeled patches are utilized to train the classifiers.}
    \label{fig:flow-chart}
\end{figure*}

Our contributions can be summarized as follows:
\begin{itemize}
    \setlength{\itemsep}{0pt}
    \item We introduce first large scale dataset of human face skin patches FaceSkin which can be used to perform classification tasks in context of deep learning.
    \item We provide benchmarks for the proposed FaceSkin using several CNNs according to each attributes.
\end{itemize}

\section{Related Work}
\subsection{Texture Dataset}
Previously, much effort has been devoted to analysing the 
usage of texture in deep learning field. DTD(Describable Textures Dataset)\cite{DTD} is an collection of textural images in the wild, annotated with a series of human-centric attributes, inspired by the perceptual properties of textures. Which has been widely applied to enrich the dataset. FMD(Flicker Material Dataset) is a database onstructed with the specific purpose of capturing a range of real world appearances of common materials\cite{FMD} Images in database are selected from Flicker manually. The materials include leather, wood, fabric, paper, etc. MINC(Material Recognition in the Wild with the Materials in Context Database)\cite{minc} is an open dataset with 23 categories, such as wood, skin, hair, sky etc. Amsterdam Library of Textures(ALOT)\cite{ALOT} is a color image collection of 250 rough textures, recorded for scientific purposes. 
\subsection{Skin Analysis in Machine Learning}
By exploring we notice that previous studies focus on skin-detection/segmentation task and the relationship between skin and diseases(cancers, lesions, etc.). Mahmoodi \textit{et al.} \cite{sdd}proposed a skin segmentation dataset with a volumn of more than 20000 images under various illumination. compaq\cite{compaq} is the first skin dataset and widely used dataset with over 10000 images. However, the quality of compaq is unsatisfied since it was published almost 20 years ago. Besides, compaq is not available any more. In terms of cancer/disease related skin analysis, The HAM10000\cite{ham10000} dataset is a large collection of multi-source dermatoscopic images of common pigmented skin lesions, which contains 10000 images  for detecting pigmented skin lesions. Yuan Liu \textit{et al} proposed a deep learning based system\cite{dls} to diagnose skin diseases. Aforementioned works proved the effectiveness and representation capacity of human skin patches. We believe pure facial skin can be leveraged in other areas.  

\section{Proposed Method}
\subsection{Face images Acquisition}
\subsubsection{Liveness Face images}

To generate high-quality facial skin patches, it is crucial to have access to face images with high resolution and a large number of subjects. However, collecting such face images in real-world scenarios can be both expensive and time-consuming. To address this challenge, we leverage the advantages offered by the public Celeb-HQ dataset as the source of our images. Celeb-HQ dataset provides the following advantages:
\begin{itemize}
    \item High resolution.
    \item Sufficient number of subjects
    \item Race coverage of White, Black, Asian, mid-estern and Latin.
\end{itemize}

\begin{figure}[htb]
    \centering
    \includegraphics[width=0.5\textwidth]{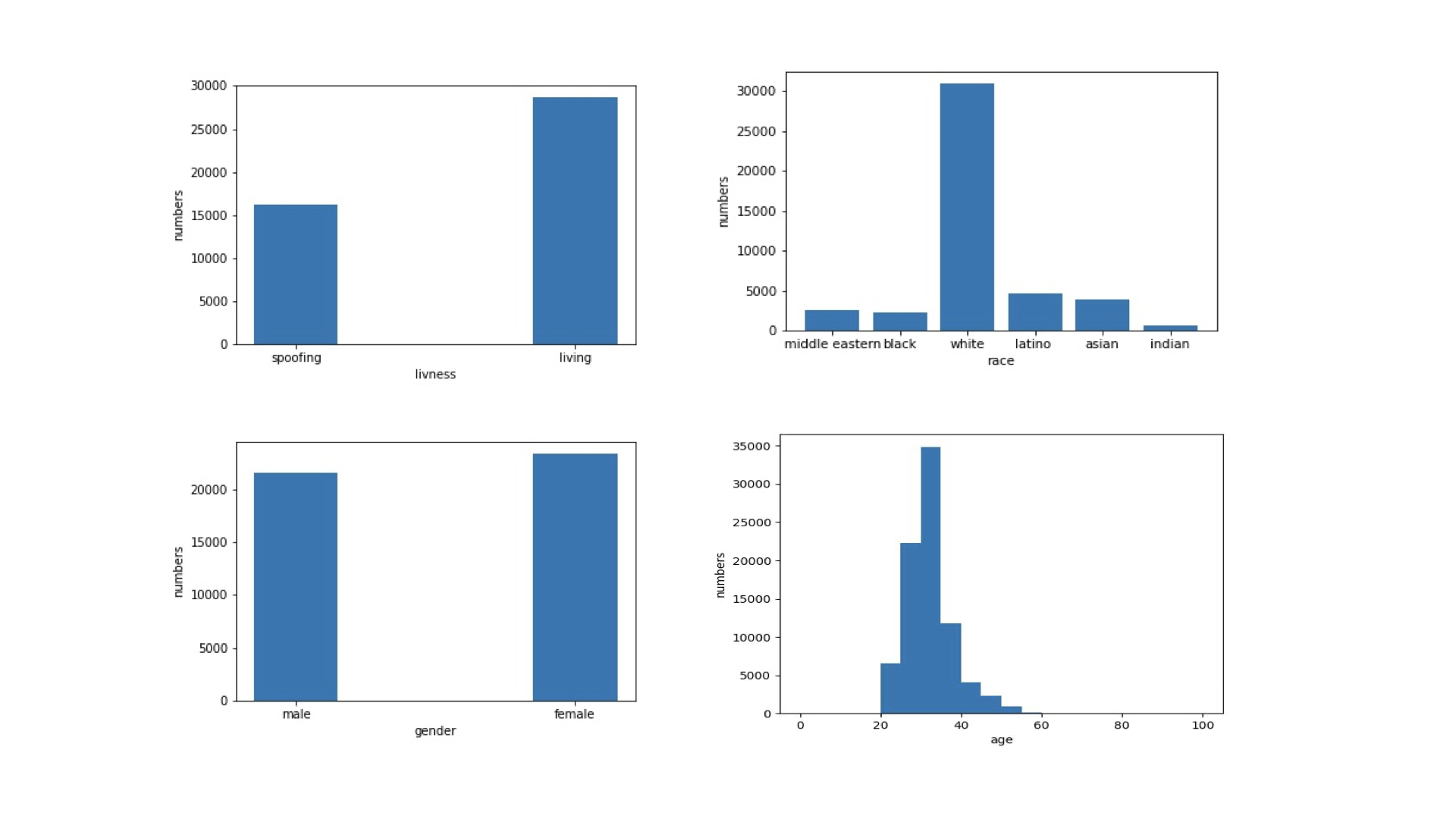 }
    \caption{From the table above, it is evident that certain attributes, such as gender and liveness, are evenly distributed within the dataset. However, attributes like race and age exhibit an uneven distribution. To address this imbalance, we take additional steps to tackle the issue and ensure a more balanced representation of these attributes in the dataset}
    \label{fig:attributes}
\end{figure}

While the Celeb-HQ images in our dataset do not come pre-labeled with the desired attributes, we leverage the facial attributes analysis module in DeepFace to obtain these attributes. DeepFace provides accurate prediction functions for facial attributes, including race, gender, and age, with accuracies of 97.44\%, 96.29\%, and 95.09\%, respectively. The predicted attributes for the CelebA-HQ dataset are visualized in Figure \ref{fig:attributes}.

However, it should be noted that certain attributes, such as race and age, may not be evenly distributed within the CelebA-HQ dataset. To address this issue, we employ a resampling strategy and supplement the dataset with additional images. Specifically, we sample images from the AFAD dataset \cite{niu2016ordinal}, which stands for Asian Face Age Dataset. To ensure high resolution, we filter out face images from AFAD that are smaller than $256\times256$ pixels and are not in frontal pose.

\subsubsection{Non-Liveness Face images}
Facial skin images possess valuable texture information that can be effectively utilized in face anti-spoofing scenarios. We select the CelebA-Spoof dataset as our source for the following reasons:

\textbf{High-Quality and Large Quantity}: The CelebA-Spoof dataset provides images of high quality and includes a significant number of samples. This ensures that we have access to a diverse range of facial images for training and evaluation purposes.

\textbf{Varied Attack Types}: CelebA-Spoof encompasses a wide variety of attack types, including print attacks, masks, paper cutouts, and replays. This diversity allows us to train and evaluate our face anti-spoofing models on various real-world attack scenarios.

\textbf{Different Dimensions}: The dataset captures images from different dimensions, including five angles, four shapes, and four camera sensors. This variability enhances the robustness and generalization of our face anti-spoofing models by exposing them to a wide range of facial variations and environmental conditions.

Considering these factors, the CelebA-Spoof dataset proves to be a suitable choice for our research and development in the face anti-spoofing domain.

\subsection{Facial skin patch Acquisition}

The skin carries valuable information for predicting individual properties. While obtaining skin patches from body parts can be challenging, these patches tend to be more stable compared to facial skin patches, which can vary due to makeup or sun exposure. With the prevalence of social media, acquiring an individual's portrait has become easy and does not raise privacy concerns.

To separate the texture (skin) from the facial structure (nose, eyes, mouth, etc.), we extract pure skin patches guided by facial landmarks. FaceMesh, provided by Mediapipe, generates detailed facial landmarks. As illustrated in Figure \ref{fig:flow-chart}, we select patches from the left and right cheeks and forehead. These regions allow us to extract pure skin patches without including additional textures from the mouth, eyes, or nose areas.

To ensure clear identification and categorization, each extracted patch is named in the format of \textit{liveness\_age\_race\_gender}, capturing the respective attributes of the skin patch.

\section{Experiment}
\subsection{Data Benchmarking}
Previous Works have demonstrated that CNN performs well in classification tasks. To verify the effectiveness of our dataset, we benchmark the dataset according to each individual attribute. We select several popular networks Resnet\cite{resnet}, Densenet\cite{densenet}, Mobilenet\cite{mobilenets} VGG\cite{vgg} and Efficientnet\cite{efficientnet} as our candidates.

\subsubsection{Age Benchmarking}
For benchmarking attribute age, we split the data into three sub-groups: under 30, 30-50 and over 50. As illustrated in \textbf{Fig} \ref{fig:attributes} , we re-sample the dataset to make an evenly distributed dataset. Only bona-fides are included. However, the cnn can't converge when fed . We propose an alternative network structure as demonstrated in \textbf{Fig.} \ref{fig:multi_path}. It has two branches, inputs of the upper branch are original portrait images, inputs of bottom branch are skin patches extracted from portrait. The features extracted from the CNN are than fused. The results show that patch-branch can enhance the representation ability compared to portrait-only approach. 

\subsubsection{Race Benchmarking}
\begin{figure}[htb]
    \centering
    \includegraphics[width=0.5\textwidth]{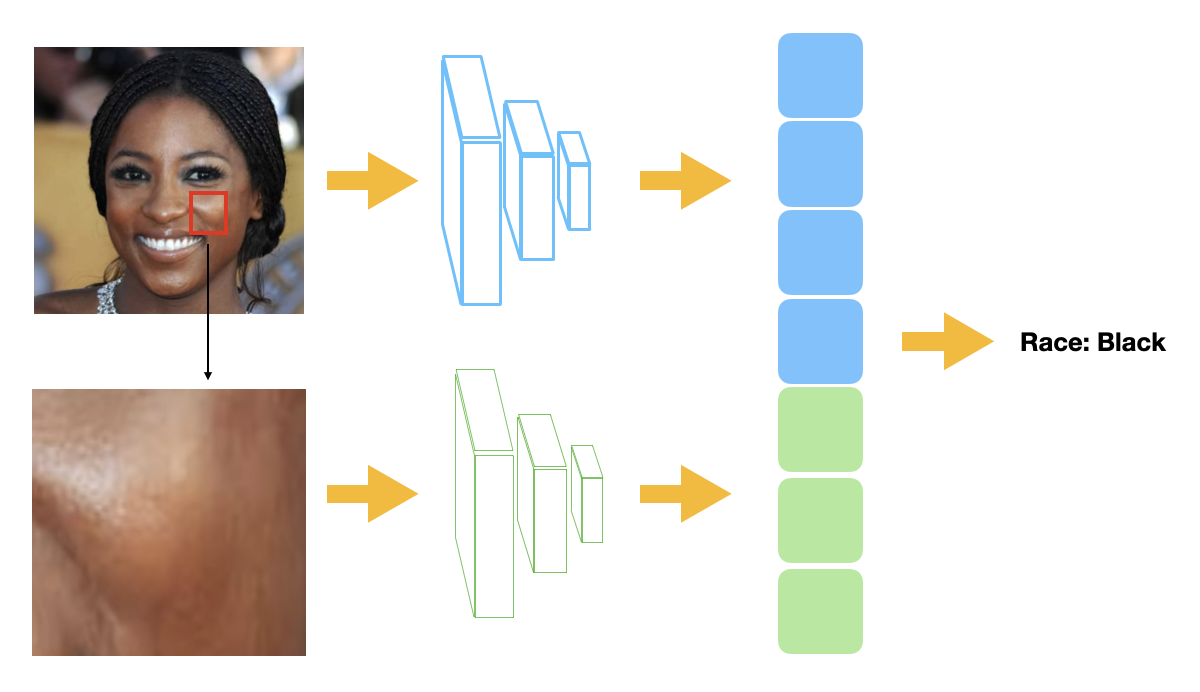 }
    \caption{Multi paths feathers fused approach. }
    \label{fig:multi_path}
\end{figure}

In Race benchmarking part, we exclude the race indian since insufficient number of images compared to other races. Through searching on communitity, no proper indian portrait datasets with high quality are found to supplement our FaceSkin. Besides, we re-sample 3000 white images to form a balanced dataset. Similiar to Age classification, the CNN can't converge given pure skin patches as inputs. We implement double-branch structure as shown in \textbf{Fig.} \ref{fig:multi_path}. The results prove the effectiveness of skin-patches.
\subsubsection{Gender Benchmarking}
As shown in \textbf{Fig.} \ref{fig:attributes}, female and male are evenly distributed, no further steps are needed to process the dataset. The results proves that pure face skin can be utilized to discriminate gender.   

\subsubsection{Liveness Benchmarking}
For liveness classification task, we utilized skin patches extracted from CelebA-spoof, which consists of both liveness and spoofing face images. The spoofing types include print attack, replay attack, 3D mask, etc. Face skin patches contain rich texture information in terms of face anti-spoofing scenario. Reflective properties of different materials make skin patches discriminative. The results indicate that facial skin patches play a vital role in face anti-spoofing field.  
        

\begin{table}[htb]
\resizebox{\linewidth}{!}{
\begin{tabular}{*{7}{c}}
  \toprule
  \multirow{2}*{CNN} & \multicolumn{2}{c}{Age} & \multicolumn{2}{c}{Race} & \multirow{2}*{Liveness} & \multirow{2}*{Gender}\\
  \cmidrule(lr){2-3}\cmidrule(lr){4-5}
  &P&P+S&P&P+S&&\\
  \midrule
    Resnet& 91.3&91.4&81.5&83.9&98.3&74.7\\
    MobileNet& 91.1&91.3&79.1&80.8&99.2&77.2\\
    DenseNet& 90.6&90.8&81.2&81.7&98.7&76.8\\
    VGG& 88.3&88.7&77.4&78.2&97.8&73.5\\
    EfficientNet& 90.8&90.9&80.6&80.9&98.4&76.3\\
  \bottomrule
\end{tabular}}
\vspace{0.5cm}
\caption{Results of experiments. P indicates portrait images while S indicates skin patches.}
\label{results}
\end{table}

\subsection{Implementation details}
Details of parameters and configurations are demonstrated in Table. 
\begin{table}[htb]
    \centering
    \begin{tabular}{|c|c|}
    \hline
     para    & value \\
    \hline
     GPU    &  Tesla T4\\
     \hline
     CPU    &  Intel Xeon 5218 \\
     \hline
     platform & Pytorch\\
     \hline
     loss function &  cross-entropy\\
     \hline
     optimizer & SGD\\
     \hline
     learning rate & 0.1\\
     \hline
     batch size & 64\\
     \hline
     epoch & 100\\
     \hline
     transform & Random Horizon filp, color jitter\\
     \hline
    \end{tabular}
    \vspace{0.5cm}
    \caption{Configurations of experiments}
    \label{tab:my_label}
\end{table}
\section{Conclusion}
In this paper, we introduce FaceSkin, a privacy-friendly dataset specifically designed for deep learning applications. We extract pure face patch images from the CelebA-HQ and Celeb-Spoof datasets. Each image in FaceSkin is carefully labeled with attributes such as age, race, liveness, and gender. 

To validate the effectiveness of the FaceSkin dataset, we conduct extensive experiments using various convolutional neural networks (CNNs). We benchmark each attribute separately and evaluate the performance of the CNN models. Additionally, for attributes that do not converge well using conventional architectures, we propose a novel two-path architecture that combines features from portrait images and facial skin patches. 

The experimental results demonstrate the effectiveness of the FaceSkin dataset and highlight its potential for attribute classification in deep learning tasks. We encourage the research community to further explore and leverage the FaceSkin dataset for advancements and applications in the field.

\bibliographystyle{plain}
\bibliography{ref}
\end{document}